\documentclass[final]{cvpr}

\usepackage{times}
\usepackage{epsfig}
\usepackage{graphicx}
\usepackage{amsmath}
\usepackage{amssymb}

\usepackage{booktabs}
\usepackage{caption}
\usepackage{scalerel}
\usepackage{array}
\usepackage{multirow}
\usepackage{dsfont}
\usepackage{stfloats}
\usepackage{url}
\usepackage{setspace}
\usepackage{booktabs}

\newcolumntype{M}{>{\centering\arraybackslash}m{.2\textwidth}}
\newcolumntype{C}[1]{>{\centering\let\newline\\\arraybackslash\hspace{0pt}}p{#1}}
\newcolumntype{R}[1]{>{\raggedleft\let\newline\\\arraybackslash\hspace{0pt}}p{#1}}
\newcolumntype{L}[1]{>{\raggedright\let\newline\\\arraybackslash\hspace{0pt}}p{#1}}

\newcommand\Bstrut{\rule[-0.9ex]{-3pt}{0pt}} 
\usepackage[pagebackref=true,breaklinks=true,colorlinks,bookmarks=false]{hyperref}

\def\cvprPaperID{280} 
\def\confYear{CVPR 2021}

\begin{document}

\title{Omni-supervised Point Cloud Segmentation via Gradual Receptive Field Component Reasoning}

\author{
Jingyu Gong\textsuperscript{\rm 1}\;\quad
Jiachen Xu\textsuperscript{\rm 1}\;\quad
Xin Tan\textsuperscript{\rm 1}\;\quad
Haichuan Song\textsuperscript{\rm 2}\;\quad\\
Yanyun Qu\textsuperscript{\rm 3}\;\quad
Yuan Xie\textsuperscript{\rm 2$*$}\;\quad
Lizhuang Ma\textsuperscript{\rm 1,2}\thanks{Corresponding Author}\\
\textsuperscript{\rm 1}Department of Computer Science and Engineering, Shanghai Jiao Tong University, Shanghai, China\\
\textsuperscript{\rm 2}School of Computer Science and Technology, East China Normal University, Shanghai, China\\
\textsuperscript{\rm 3}School of Informatics, Xiamen University, Fujian, China\\
{\tt\small \{gongjingyu,xujiachen,tanxin2017\}@sjtu.edu.cn\;\quad  hcsong@cs.ecnu.edu.cn}\\
{\tt\small yyqu@xmu.edu.cn\;\quad yxie@cs.ecnu.edu.cn\;\quad ma-lz@cs.sjtu.edu.cn}
}

\maketitle


\begin{abstract}
   Hidden features in neural network usually fail to learn informative representation for 3D segmentation as supervisions are only given on output prediction, while this can be solved by omni-scale supervision on intermediate layers. In this paper, we bring the first omni-scale supervision method to point cloud segmentation via the proposed gradual Receptive Field Component Reasoning (RFCR), where target Receptive Field Component Codes (RFCCs) are designed to record categories within receptive fields for hidden units in the encoder. Then, target RFCCs will supervise the decoder to gradually infer the RFCCs in a coarse-to-fine categories reasoning manner, and finally obtain the semantic labels. Because many hidden features are inactive with tiny magnitude and make minor contributions to RFCC prediction, we propose a Feature Densification with a centrifugal potential to obtain more unambiguous features, and it is in effect equivalent to entropy regularization over features. More active features can further unleash the potential of our omni-supervision method. 
   We embed our method into four prevailing backbones and test on three challenging benchmarks. Our method can significantly improve the backbones in all three datasets. Specifically, our method brings new state-of-the-art performances for S3DIS as well as Semantic3D and ranks the 1st in the ScanNet benchmark among all the point-based methods. 
   Code will be publicly available at https://github.com/azuki-miho/RFCR. 
   
\end{abstract}

\section{Introduction}
Semantic segmentation of point cloud in which we need to infer the point-level labels is a typical but still challenging task in 3D vision. Meanwhile, this technique can be widely used in many applications like robotics, autonomous driving, and virtual/augmented reality.

\begin{figure}
    \centering
    \includegraphics[width=\linewidth]{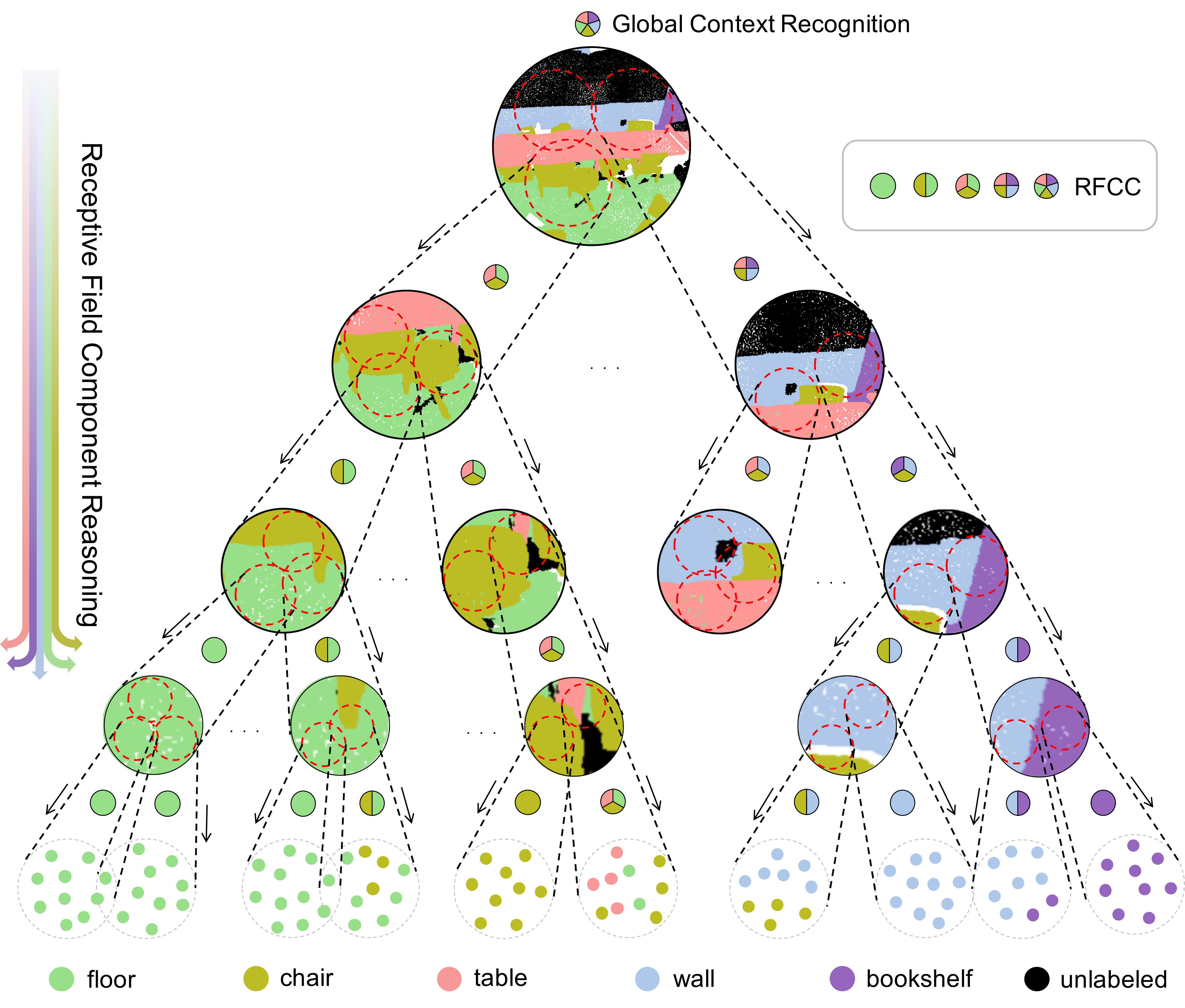}
    \caption{Illustration of Receptive Field Component Reasoning for a point cloud in ScanNet v2 from top to bottom. The Receptive Field Component Code (RFCC) indicates the category components in the receptive field. In the decoding stage, the segmentation problem is decomposed into a much easier global context recognition problem (predicting the global RFCCs, see the top of figure) and a series of receptive field component reasoning problems. During reasoning, the target RFCCs generated in the encoder are used as the groundtruth in the decoder to guide the network to gradually reason the RFCCs in a coarse-to-fine manner, and finally obtain the semantic labels.
    }
    \label{fig:reasoning}
\end{figure}

To handle point cloud segmentation, previous works usually introduced well-designed encoder-decoder architecture to hierarchically extract global context features in the encoding stage, and distribute contextual features to points in the decoding stage to achieve point-wise labeling~\cite{graham20183d, thomas2019kpconv, yan2020pointasnl}. However, in the typical encoder-decoder framework, network is merely supervised by labels of points in the final layer~\cite{wu2019pointconv, thomas2019kpconv, hu2020randla}, while ignoring a critical fact that, hidden units in other layers lack direct supervision to extract features with informative representation. In other words, multi-scale/omni-scale supervision is indeed necessary.

In 2D vision, CVAE~\cite{sohn2015learning} attempted to give a multi-scale prediction and supervision to extract useful features in segmentation task. CPM~\cite{wei2016convolutional} and MSS-net~\cite{ke2018multi} tried to add intermediate supervision periodically and layer-wise loss, respectively. PointRend~\cite{kirillov2020pointrend} proposed to segment image in low-resolution, and iteratively up-sample the coarse prediction and fine-tune it to obtain final result, thus prediction at different scales can be supervised together.

However, so far, no one succeed in applying multi-scale, let alone omni-scale supervision to 3D semantic segmentation, due to the irregularity of point cloud. Unlike in image domain, it is hard to up-sample the hidden features to the original resolution through simple tiling or interpolation, because there is no fixed mapping relationship between sampled point cloud and original point cloud especially when the sampling is random~\cite{wu2019pointconv,hu2020randla}. 
Additionally, the common up-sampling methods using nearest neighbors cannot trace the encoding relationship, thus introducing improper supervisions to the intermediate features (referring Sec~\ref{subsec:ablation} for discussion).
More recently, SceneEncoder~\cite{xu2020sceneencoder} provided a method to supervise the center-most layer to extract meaningful global features, but lots of other layers remain unhandled.


To solve this problem, we propose an omni-scale supervision method via gradual Receptive Field Component Reasoning. Instead of up-sampling the hidden features to the original resolution, we design a Receptive Field Component Code (RFCC) to effectively trace the encoding relationship and represent the categories within receptive field for each hidden unit. Based upon this, we generate the target RFCCs at different layers from semantic labels in the encoding stage to supervise the network at all scales. Specifically, in the decoding stage, the target RFCCs will supervise the network to predict the RFCCs at different scales, and the features (hints) from skip link can help further deduce RFCCs within more local and specific receptive fields. In this way, the decoding stage is transferred into a gradual reasoning procedure, as shown in Figure~\ref{fig:reasoning}.

Inspired by SceneEncoder~\cite{xu2020sceneencoder}, for each sampled point in any layer of encoder, according to the existence of categories in its receptive field, a multi-hot binary code can be built, designated as target Receptive Field Component Code (RFCC). The target RFCCs at different layers are generated alongside the convolution and down-sampling, thus they can precisely record the existing categories in corresponding receptive fields without any extra annotations. In Figure~\ref{fig:reasoning}, we show the target RFCCs at various layers for a point cloud in the decoding stage, where the network will first recognize the global context (inferring the categories of objects existing in the whole point cloud). Then, contextual features will be up-sampled iteratively to gradually reason the RFCCs in a coarse-to-fine manner. By comparing the target RFCCs and the predicted RFCCs, the omni-scale supervision can be realized. It is noteworthy that even the network reasons the RFCCs gradually, the training and inference of network is implemented in a end-to-end manner.

Additionally, to further unleash the potential of omni-scale supervision, more active features (features with large magnitude) are required to make unambiguous contribution to the RFCC prediction. Contrarily, in traditional networks~\cite{wu2019pointconv,thomas2019kpconv,xu2020sceneencoder}, lots of units are inactive with tiny magnitude, such that having minor contribution to the final prediction. The principle underlying the above observations comes from entropy regularization~\cite{grandvalet2005semi,lee2013pseudo} over features, where greater number of active dimensionalities would bring low-density separation between positive features and negative features, generating more unambiguous features with certain signals. Consequently, in point cloud scenario, more certainty in features can help the training of the network to better reason the RFCCs at various scales and finally predict the semantic labels. Motivated by this, we proposed a Feature Densification method with a well-deigned potential function to push hidden features away from $0$. Moreover, this potential is in effect equivalent to a entropy loss over features (detailed deduction is shown in Sec~\ref{subsec:fd}), leading to a simple but highly effective regularization for intermediate features.

To evaluate the performance and versatility of our method in point cloud semantic segmentation task, we embed our method into four prevailing backbones (deformable KPConv, rigid KPConv~\cite{thomas2019kpconv}, RandLA~\cite{hu2020randla}, and SceneEncoder~\cite{xu2020sceneencoder}), and test on three challenging point cloud datasets (ScanNet v2~\cite{dai2017scannet} for indoor cluttered rooms, S3DIS~\cite{armeni20163d} for large indoor space, and Semantic3D~\cite{hackel2017semantic3d} for large-scale outdoor space). In all the three datasets, we outperform the backbone methods and almost all the state-of-the-art point-based competitors. What's more, we also push the state-of-the-art of S3DIS~\cite{armeni20163d} and Semantic3D~\cite{hackel2017semantic3d} ahead.

\section{Related Work}

\paragraph{Point Cloud Semantic Segmentation.}
PointNet~\cite{qi2017pointnet} proposed to directly concatenate global features to point-wise features before several Multi-Layer Perceptrons (MLPs) to finish the semantic segmentation. Later, PointNet++~\cite{qi2017pointnet++}, SubSparseConv~\cite{graham20183d} and KPConv~\cite{thomas2019kpconv} utilized an encoder-decoder architecture with skip links for better fusion of local and global information. Joint tasks like instance segmentation and edge detection are also introduced to enhance the performance of semantic segmentation through additional supervision~\cite{pham2019jsis3d, zhao2020jsnet, hu2020jsenet}. SceneEncoder~\cite{xu2020sceneencoder} designed a meaningful global scene descriptor to guide the global feature extraction. These methods directly utilized semantic labels to supervise the output features or features in the center-most layer.

Compared with previous works, we propose an omni-scale supervision method for point cloud semantic segmentation via a gradual Receptive Field Component Reasoning.

\paragraph{Multi-scale Supervision.} 
In 2D Vision, CVAE~\cite{sohn2015learning} proposed to give multi-scale prediction in the segmentation task. RMI~\cite{zhao2019region} proposed to predict and supervise the neighborhood of each pixel rather than the pixel itself. PointRend~\cite{kirillov2020pointrend} segmented the images in a coarse-to-fine fashion, \ie give low-resolution prediction, and iteratively up-sample and fine-tune it to obtain the original-resolution prediction. CPM~\cite{wei2016convolutional} and MSS-net~\cite{ke2018multi} added intermediate supervision periodically and layer-wise loss, respectively.


Compared with these methods, we design a Receptive Field Component Code (RFCC) to represent receptive field component and dynamically generate target RFCCs to give omni-scale supervision to the network rather than simply up-sample the features to the original resolution or down-sample the ground truth. Thanks to the omni-scale supervision, the network can infer the RFCCs gradually and finally obtain RFCCs in the original resolution which is also the semantic labels.

\paragraph{Entropy Regularization.}
Entropy Regularization~\cite{grandvalet2005semi} minimized the prediction entropy in semi-supervised classification task to obtain unambiguous final features. This idea is introduced into the deep neural network for self-training by \cite{lee2013pseudo}, and the final features with tiny magnitude will be pushed away from $0$ to make deterministic contribution to the final prediction. In these methods, final features with positive values will be greater and negative features will be smaller due to the entropy loss. 

Compared with their methods, our Feature Densification introduce the entropy regularization~\cite{grandvalet2005semi,lee2013pseudo} into the hidden features rather than just the final features to obtain more active hidden features which can directly contribute to the RFCC prediction.

\section{Methods}
\label{sec:methods}

\begin{figure*}[th]
    \centering
    \includegraphics[width=0.9\linewidth]{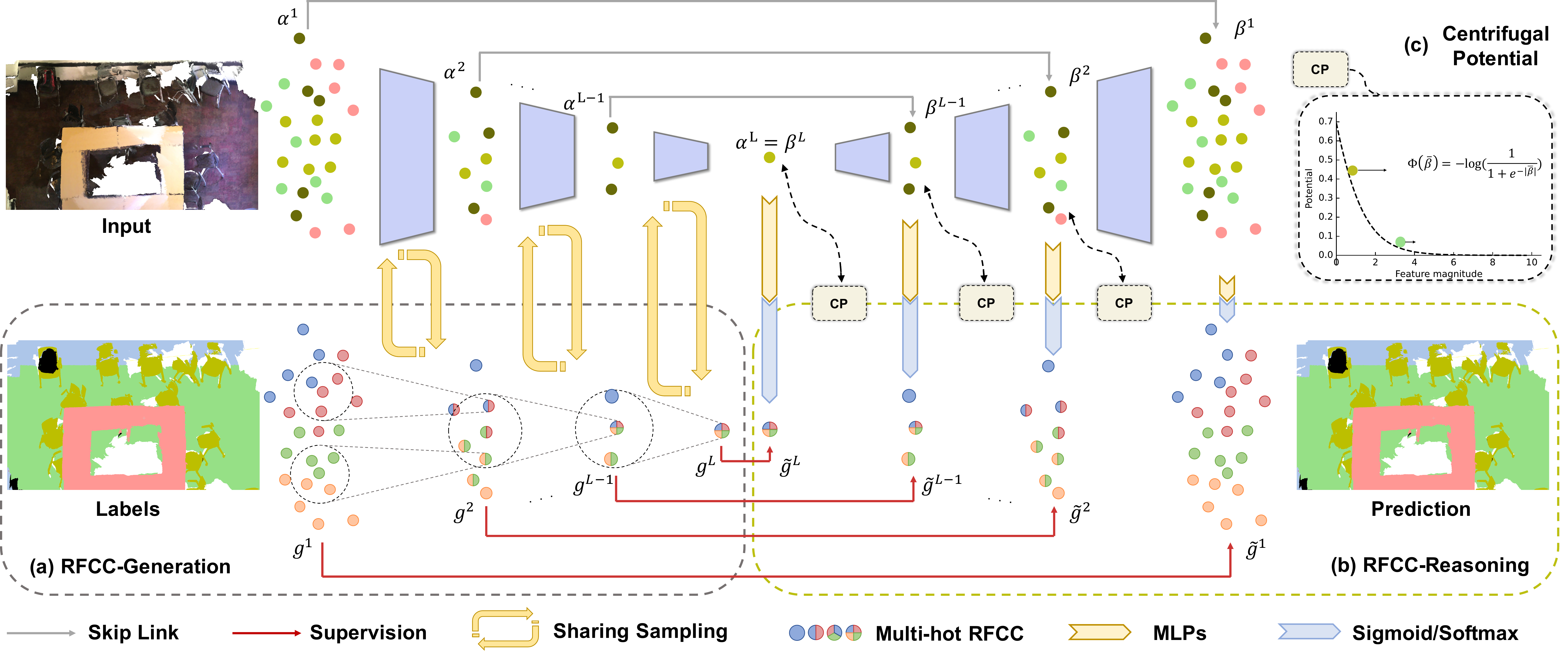}
    \caption{Framework of gradual Receptive Field Component Reasoning. (a) shows the target Receptive Field Component Codes (RFCCs) is generated alongside the common encoding procedure. (b) indicates the network will predict the RFCCs in a coarse-to-fine manner. (c) represents the centrifugal potential which pushes hidden features away from $0$. In our network, the target RFCCs will supervise the RFCC predictions, and the learnt feature can reason RFCCs in more local and specific receptive fields as more and more local features (clues) are provided through skip links. The prediction activation function will be Softmax for the final layer and Sigmoid otherwise.}
    \label{fig:framework}
\end{figure*}

In the following parts, we will first give an overview of our method in Sec~\ref{subsec:overview}. Then, we will introduce the Receptive Field Component Codes (RFCCs) and the target RFCCs that we generate at various layers in Sec \ref{subsec:rfcc}. In Sec \ref{subsec:reasoning}, how to use these target RFCCs to supervise the network, and make the gradual Receptive Field Component Reasoning, would be explained. At last, we will show the strategy of Feature Densification for more active features in Sec \ref{subsec:fd}.

\subsection{Overview}
\label{subsec:overview}

The framework of our gradual Receptive Field Component Reasoning (RFCR) is shown in Figure~\ref{fig:framework}. In our method, we generate target Receptive Field Component Codes (RFCCs) at different layers alongside the convolution and sampling of features (Figure~\ref{fig:framework} (a)) in the encoding stage. In the decoding stage, the network will reason the RFCCs at different layers, and the corresponding target RFCCs will give omni-scale supervision on the predicted RFCCs (Figure~\ref{fig:framework} (b)). Consequently, the semantic segmentation task can be treated as a coarse-to-fine receptive field component reasoning procedure after recognizing the global context (predicting categories of objects existing in the point cloud). Additionally, we introduce Feature Densification through a centrifugal potential to obtain more active features for omni-scale RFCC prediction (Figure~\ref{fig:framework} (c)).

\subsection{Receptive Field Component Code}
\label{subsec:rfcc}
For a point cloud, it is easy to define the label of a point in the original point cloud. Nevertheless, it is non-trivial to give a label to a point in any down-sampled point cloud which receives information from points inside its receptive field.
In our method, we design a Receptive Field Component Code (RFCC) to represent all categories within the receptive field of sampled points in the encoder. The target RFCCs are generated alongside the convolution and sampling of features in the encoding stage. In other words, sharing sampling is used between the encoding stage (left part of top branch in Figure~\ref{fig:framework}) and RFCC generation (Figure~\ref{fig:framework} (a)), thus the generated target RFCCs can precisely record the category components in the receptive fields, even though the sampling of point cloud is a random process.

\paragraph{Implementation.} Our RFCC is designed to be a multi-hot label for every point in any layer of encoder. Specifically, in the semantic segmentation task where we need to classify each point into $C$ categories, the RFCC will be a $1\times C$ binary vector. Given the $i$-th point in the $l$-th layer of the encoder $p_i^l$, the target RFCC $g_i^l$ represents the categories of objects existing in the receptive field of $p_i^l$, and each element $g_i^l[k]$ indicates the existence of category $k$.
Based upon this definition, we can first assign the one-hot label of input point $p_i$ to the RFCC $g_i^1$ in the input layer, because the receptive field of point $p_i$ only contains $p_i$ itself:
\begin{equation}
    g_i^1 = \text{one-hot}(y_i),
\end{equation}
where $y_i$ is the label of point $p_i$ in the original point cloud.
As illustrated in Figure~\ref{fig:framework} (a), we can obtain $g_i^l$ from the RFCCs in the previous layer $g_i^{l-1}$ alongside the 3D Convs:
\begin{equation}
    g_i^l[k] = \mathop{\lor}\limits_{j\in\mathcal{N}(i)} \{g_j^{l-1}[k]\}
\end{equation}
where $k\in [1,C]$ indicates the channel index, and $j$ is the index of point in $p_i^l$'s receptive field at the $(l-1)$-th layer. That is to say, $p_i^l$ receives features from $p_j^{l-1}$ in the 3D Convs thanks to the sharing sampling. $\lor$ represents the logical OR (disjunction) operation. It is noteworthy that the generation of RFCCs only occurs in the encoder, rather than the decoder. The generation of RFCCs is iterated until reaching the center-most layer $L$.
Typically, the scene descriptor is only a naturally deduced global supervisor when the center-most layer contains only one point~\cite{xu2020sceneencoder}. Besides, $g_i^2$ can also be treated as a simplified version of neighborhood multi-dimension distribution in RMI~\cite{zhao2019region}, which exploits the semantic relationship among neighboring points.

\subsection{RFCC Reasoning}
\label{subsec:reasoning}
The decoder of network is to infer the category of each input point in the task of semantic segmentation. In our method, as shown in Figure~\ref{fig:framework} (b), we decompose this complex problem into a much easier global context recognition problem (predicting $g_i^L$) and a series of gradual receptive field component reasoning problem (reasoning $g_i^{l-1}$ from $g_i^l$ gradually with additional features $\alpha_i^l$ from skip link and finally obtain the semantic labels $g_i^1$).


As shown in Figure~\ref{fig:framework}, ${\beta}_i^l$ is the features of sampled point $p_i^l$ in decoder. For each layer of decoder except the last one, we apply a shared Multi-Layer Perceptron (MLP) $\mathcal{M}^l$ and a sigmoid function $\sigma$ to ${\beta}_i^l$ to predict the RFCCs $\tilde{g}_i^l$:
\begin{equation}
    \label{eq:prediction}
    \tilde{g}_i^l = \sigma(\mathcal{M}^l({\beta}_i^l)).
\end{equation}
Then, the target RFCC $g_i^l$ generated in the encoding stage is directly used to guide $\tilde{g}_i^l$ prediction through layer-wise supervision $\mathcal{L}_R^l$:
\begin{equation}
    \mathcal{L}_R^l = -\frac{1}{C|P^l|}\sum_{i=1}^{|P^l|}\sum_{k=1}^C \mathcal{L}_R^l(i,k) ,
\end{equation}
where
\begin{equation}
    \mathcal{L}_R^l(i,k) = g_i^l[k]log(\tilde{g}_i^l[k])+ (1-g_i^l[k])log(1-\tilde{g}_i^l[k]),
\end{equation}
$P^l$ denotes the sampled point cloud in the $l$-th layer of encoder, and $|P^l|$ corresponds the number of points in $P^l$.

According to Eq.~(\ref{eq:prediction}), the center-most features $\beta_i^L$ which contain global information will learn to recognize the global context, \ie, predict $\tilde{g}_i^L$ with largest receptive field. Meanwhile, $g_i^L$ will be used to regularize this prediction to help $\beta_i^L$ learn a better representation. Then, for the following layer of decoder, $\beta^L$ which learns informative representation to predict $\tilde{g}_i^{L}$ will be up-sampled and concatenated with ${\alpha}_i^{L-1}$ from the skip link. After that, the concatenated features will be used to extract more distinguishable $\beta_i^{L-1}$ via 3D Convs, and the extracted features $\beta_i^{L-1}$ will be used to reason the RFCCs $\tilde{g}_i^{L-1}$ of more local and specific receptive field. This procedure is iterated until $l=2$. The whole RFCC reasoning loss can be simply expressed by
\begin{equation}
    \mathcal{L}_R = \frac{1}{L-1}\sum_{l=2}^L \mathcal{L}_R^l.
\end{equation}
In the last layer, we can simply utilize the MLPs and softmax to predict the $\tilde{g}_i^1$, and cross entropy loss is used to supervise the output features in the original scale.

\subsection{Feature Densification}
\label{subsec:fd}
Due to the large amounts of supervision introduced by the gradual Receptive Field Component Reasoning, more active features with unambiguous signals are required. However, there are many inactive hidden units with tiny magnitude in the traditional network (detailed experiment is shown in Sec~\ref{subsec:ablation}). Therefore, we introduce a centrifugal potential to bring low-density separation between positive features and negative features (\ie push features away from $0$) as shown in Figure~\ref{fig:framework} (c):
\begin{equation}
    \Phi(\bar{\beta}) = -\text{log}\frac{1}{1+e^{-\vert \bar{\beta} \vert}},
\end{equation}
where $\bar{\beta}=a(\beta)$ and $a$ can be an identity function or a simple perceptron. We can see the negative gradient of potential function over feature is:
\begin{equation}
\label{eq:gradient}
    -\frac{\partial \Phi(\bar{\beta})}{\partial \bar{\beta}} = \text{sign}(\bar{\beta})\frac{e^{-\vert \bar{\beta} \vert}}{1 + e^{-\vert \bar{\beta} \vert}}
\end{equation}
which have the same sign as the feature. This indicates positive features will become greater and negative features will be smaller given this potential. Additionally, features with smaller absolute value will receive larger gradient according to this formula.

Meanwhile, this centrifugal potential can be implemented by a simple entropy loss:
\begin{equation}
\label{eq:fd_feature_loss}
    \begin{array}{rcl}
    \mathcal{L}_F^l(i,k) & = & \Phi(\bar{\beta}_{i,k}^l)\\
    & = & -\text{log}\frac{1}{(1+e^{-\vert \bar{\beta}_{i,k}^l\vert})}\\
    & = & \left\{
    \begin{array}{ll}
    -\text{log}(\sigma(\bar{\beta}_{i,k}^l)) & \bar{\beta}_{i,k}^l \ge 0\\
    -\text{log}(1-\sigma(\bar{\beta}_{i,k}^l)) & \bar{\beta}_{i,k}^l < 0
    \end{array}
    \right. ,\\
    \end{array}
\end{equation}
where $\bar{\beta}_{i,k}^l$ is the $k$-th channel of $\bar{\beta}_i^l$.

If we take the following notation:
\begin{equation}
    \begin{array}{l}
    \tilde{t}_{i,k}^l = \sigma(\bar{\beta}_{i,k}^l)\\
    t_{i,k}^l = 
    1 ~\text{if}~\bar{\beta}_{i,k}^l \ge 0,~0 ~\text{if}~\bar{\beta}_{i,k}^l < 0 ,
    \end{array}
\end{equation}
we can reformulate Eq.~(\ref{eq:fd_feature_loss}) into
\begin{equation}
    \mathcal{L}_F^l(i,k) = -[t_{i,k}^l\text{log}(\tilde{t}_{i,k}^l) + (1-t_{i,k}^l)\text{log}(1-\tilde{t}_{i,k}^l)].
\end{equation}

So, our centrifugal potential can be treated as entropy regularization~\cite{lee2013pseudo} over hidden features which can decrease ambiguity of features in the intermediate layers. On the other side, our omni-scale supervision can directly benefit from more active features with certain signal introduced by the Feature Densification. That is because more unambiguous features can participate into the RFCC predictions and help learning better representation of hidden layer, improving the semantic segmentation performance.

The total loss for Feature Densification can be summarized by
\begin{equation}
\label{eq:fd_loss}
    \mathcal{L}_F = \frac{1}{L-1}\sum_{l=2}^L\frac{1}{|P^l|K^l}\sum_{i=1}^{|P^l|}\sum_{k=1}^{K^l} \mathcal{L}_F^l(i,k),
\end{equation}
and $K^l$ represents the number of features' channel in $\bar{\beta}_i^l$.

In a nutshell, all the supervision can be concluded by 
\begin{equation}
    \mathcal{L} = \mathcal{L}_S + \lambda_1 \mathcal{L}_R + \lambda_2 \mathcal{L}_F.
\end{equation}
where $\lambda_1$ and $\lambda_2$ are two adjustable hyper-parameters while $\mathcal{L}_S$ represents the common cross entropy loss for semantic segmentation. In our experiment, we simply set $\lambda_1$ and $\lambda_2$ to $1$, and we find it can perform well in most cases.

\section{Experiments}
\label{sec:exp}
To show the effectiveness of our method and prove our claims, we embed our method into four prevailing methods (deformable KPConv, rigid KPConv~\cite{thomas2019kpconv}, RandLA~\cite{hu2020randla} and SceneEncoder~\cite{xu2020sceneencoder}), and conduct experiments on three popular point cloud segmentation datasets (ScanNet v2~\cite{dai2017scannet} for cluttered indoor scenes, S3DIS~\cite{armeni20163d} for large-scale indoor rooms and Semantic3D~\cite{hackel2017semantic3d} for large outdoor spaces). First, we introduce these three datasets in Sec \ref{subsec:datasets}. Next, implementation details and hyper-parameters used in our experiments are described in Sec \ref{subsec:implementation}. Then, we give the metric used to evaluate the performance as well as the quantitative and qualitative results in Sec \ref{subsec:result}. Finally, we conduct more ablation studies to prove our claims in Sec \ref{subsec:ablation}.

\subsection{Datasets}
\label{subsec:datasets}
\paragraph{ScanNet v2.} In the task of ScanNet v2~\cite{dai2017scannet}, we need to classify all the points into $20$ different semantic categories. This dataset provides $1,513$ scanned scenes with point-level annotations, $1,201$ scanned scenes for training, and $312$ scanned scenes for validation. Another $100$ scanned scenes are published without any annotations for testing. We need to make prediction on the test set and submit our final result to ScanNet server for testing. 

\paragraph{S3DIS.} S3DIS~\cite{armeni20163d} provides point clouds of $271$ rooms with comprehensive annotations in $6$ large-scale indoor areas from $3$ different buildings. There are $273$ million points in total, and all these points are categorized into $13$ classes.  Following~\cite{qi2017pointnet, thomas2019kpconv}, we take Area 5 as the test set and rooms in the remaining areas for training.

\paragraph{Semantic3D.} Semantic3D~\cite{hackel2017semantic3d} is a large-scale outdoor point cloud dataset with online benchmark. It contains more than $4$ billion points from diverse urban scenes, and all the points are classified into $8$ categories. The whole dataset includes $15$ point clouds for training and another $15$ point clouds for testing. For easy evaluation, Semantic3D provides the task of Semantic3D reduced-8, where $15$ large-scale point clouds are used for training and $4$ down-sampled point clouds are used for testing. 

\subsection{Implementation}
\label{subsec:implementation}
All the experiments can be conducted on a single GTX 1080Ti with 3700X CPU and 64 GB RAM. We apply our method to a common backbone deformable KPConv~\cite{thomas2019kpconv} and evaluate the performance on all three datasets. To show the versatility of our method, we also embed our method into three other backbones (one for each dataset).
\paragraph{ScanNet.}
We separately choose deformable KPConv~\cite{thomas2019kpconv} and SceneEncoder~\cite{xu2020sceneencoder} as our backbones and apply our method.
When we take deformable KPConv as our backbone, we randomly sample spheres with radius equal to $2$ meters from scenes in the training set during training procedure, and the batch size is set to $10$.
When we take SceneEncoder as our backbone and train our model, we randomly sample $8$ $3$m$\times$$1.5$m$\times$$1.5$m cubes from training scenes for every batch like SceneEncoder~\cite{xu2020sceneencoder}. 
After training, we separately predict the results of the test set using these two trained models and submit them to the online benchmark server for testing~\cite{dai2017scannet}. 

\paragraph{S3DIS.}
We insert our methods into deformable KPConv~\cite{thomas2019kpconv} and RandLA~\cite{hu2020randla} respectively and treat them as our backbones. When we take deformable KPConv as our backbone, we randomly sample spheres with $2$m radius from original point clouds, and the batch size is set to $5$. We randomly sample $40,960$ points from entire rooms for each training sample and set the batch size to be $6$ when taking RandLA~\cite{hu2020randla} as the backbone. Rooms in Area-1,2,3,4,6 are used for training. After training, we test the model on the whole S3DIS Area-5 set.

\paragraph{Semantic3D.} Deformable KPConv and rigid KPConv proposed in~\cite{thomas2019kpconv} are taken as our backbones to evaluate our method on Semantic3D reduced-8 task~\cite{hackel2017semantic3d}. Because Semantic3D is a large-scale outdoor space dataset, point cloud is randomly sampled into a sphere with $3$m radius for deformable KPConv backbone and $4$m radius for rigid KPConv backbone. Every time, $10$ samples are fed into the network for training and testing. We need to submit the final predictions to the Semantic3D server for testing~\cite{hackel2017semantic3d}.

\subsection{Metric and Results}
\label{subsec:result}
\paragraph{Metric.} For better evaluation of segmentation performance, we take mean Intersection over Union (mIoU) among categories as our metric like many previous works~\cite{gong2021boundary, qi2017pointnet, thomas2019kpconv}.

\begin{figure}[th]
    \centering
    \includegraphics[width=\linewidth,height=4.6cm]{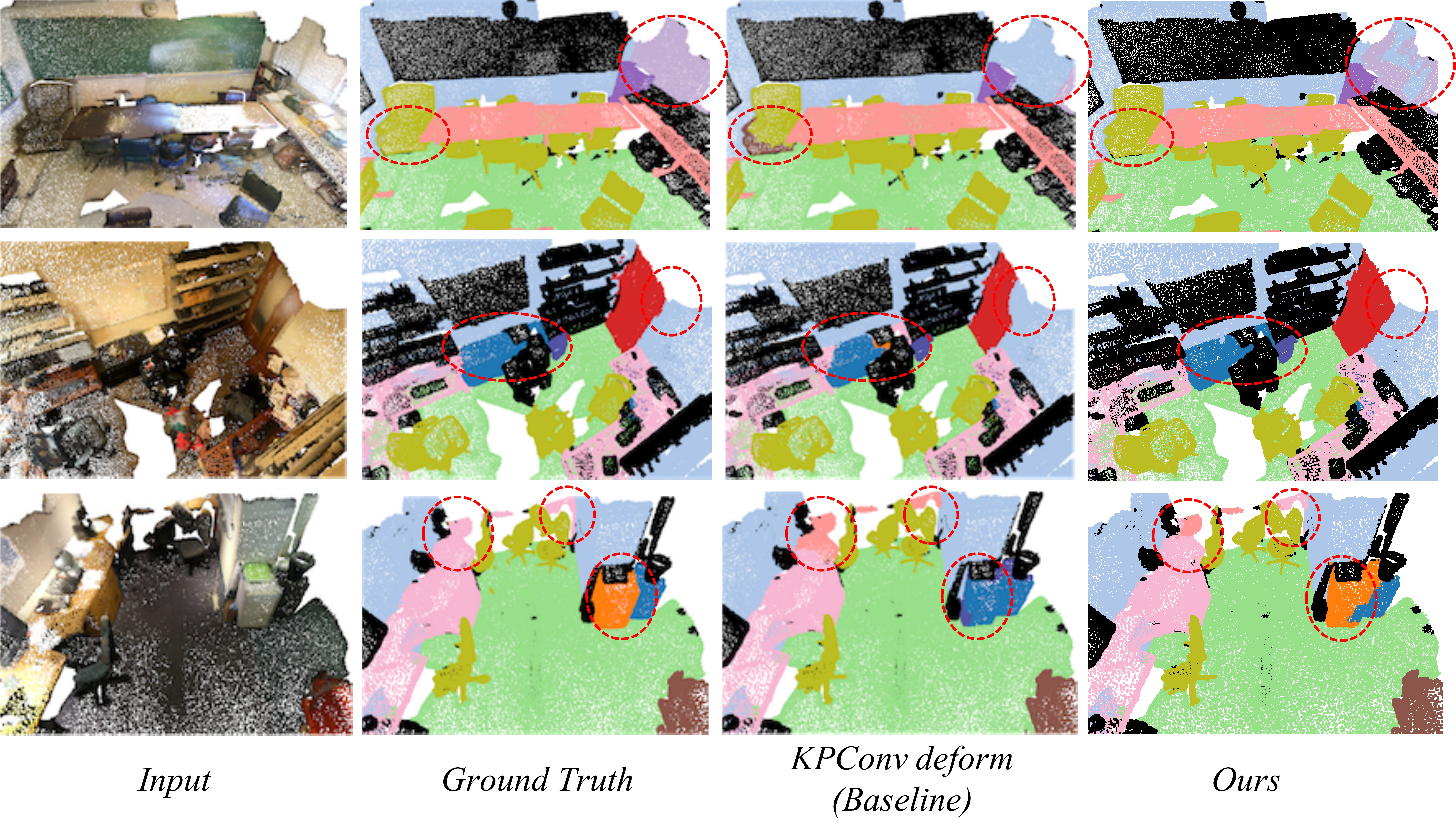}
    \caption{Visualization results on the validation dataset of ScanNet v2. The images from the left to right are input point clouds, semantic labels, predictions given by our baseline and our method, respectively.}
    \label{fig:scannet}
\end{figure}

The results of semantic segmentation on ScanNet v2~\cite{dai2017scannet} are reported in Table~\ref{tab:scannet}, where we achieve $70.2\%$ mIoU and rank first in this benchmark among all point-based methods. Here, we take deformable KPConv as our baseline and $1.8\%$ improvement is achieved in mIoU. To show the generalization ability of our method, we also apply our method to SceneEncoder~\cite{xu2020sceneencoder}. As shown in Table~\ref{tab:scannet}, $3.1\%$ improvement in mIoU is achieved. 
Additionally, we provide the qualitative results of our baseline (deformable KPConv) and our method in Figure~\ref{fig:scannet}. The red dashed circles indicate the obvious qualitative improvements.

\begin{table}
\centering
\begin{tabular}{lccc}  
\toprule
Method  & mIoU(\%) \\
\midrule
PointNet++ ({\color{blue}NIPS'17})~\cite{qi2017pointnet++} & 33.9 \\
PointCNN ({\color{blue}NIPS'18})~\cite{li2018pointcnn} & 45.8 \\
3DMV ({\color{blue}ECCV'18})~\cite{dai20183dmv} &48.4 \\ 
PointConv ({\color{blue}CVPR'19})~\cite{wu2019pointconv} & 55.6\\
TextureNet ({\color{blue}CVPR'19})~\cite{huang2019texturenet} & 56.6 \\
HPEIN ({\color{blue}ICCV'19})~\cite{jiang2019hierarchical} &61.8\\
SPH3D-GCN ({\color{blue}TPAMI'20})~\cite{lei2020spherical} & 61.0\\
FusionAwareConv ({\color{blue}CVPR'20})~\cite{zhang2020fusion} & 63.0 \\ 
FPConv ({\color{blue}CVPR'20})~\cite{lin2020fpconv} & 63.9\\
DCM-Net ({\color{blue}CVPR'20})~\cite{Schult_2020_CVPR} & 65.8 \\
PointASNL ({\color{blue}CVPR'20})~\cite{yan2020pointasnl} & 66.6 \\
FusionNet ({\color{blue}ECCV'20})~\cite{zhang2020deep} & 68.8\\
\midrule
SceneEncoder ({\color{blue}IJCAI'20})~\cite{xu2020sceneencoder} & 62.8\\
SceneEncoder + Ours & 65.9\\
\midrule
KPConv \textit{deform} ({\color{blue}ICCV'19})~\cite{thomas2019kpconv} & 68.4 \\
KPConv \textit{deform} + Ours & \textbf{70.2}\\
\bottomrule
\end{tabular}
\caption{Results of indoor scene semantic segmentation segmentation on ScanNet v2.}
\label{tab:scannet}
\end{table}

\begin{figure}[ht]
    \centering
    \includegraphics[width=\linewidth,height=4.6cm]{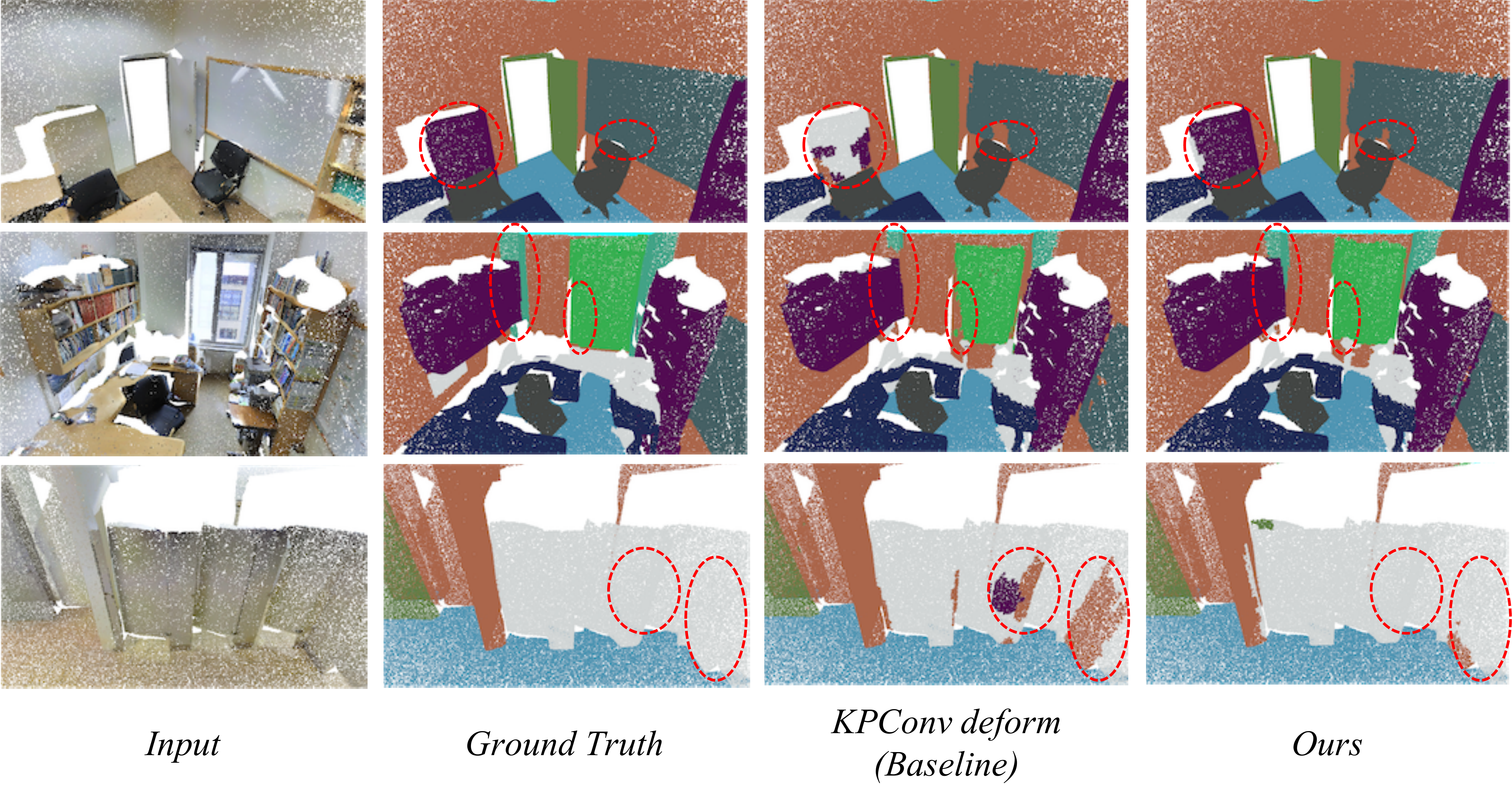}
    \caption{Visualization results on the test dataset of the S3DIS Area-5. The left-most images are input point clouds and the following images are segmentation ground truth, predictions of baseline and our method separately.}
    \label{fig:s3dis}
\end{figure}

We report the segmentation results on S3DIS Area-5~\cite{armeni20163d} in Table~\ref{tab:s3dis}. In this dataset, we also take deformable KPConv as our backbone and achieve $68.73\%$ mIoU in S3DIS Area-5 task which pushes the state-of-the-art performance ahead. Deformable KPConv is also treated as our baseline for its good performance. Meanwhile, we also apply our method to RandLA and the improvement over these backbones is also obvious (i.e., $2.67\%$ mIoU). Figure~\ref{fig:s3dis} gives the visualization results of our method and the qualitative improvement over the baseline (deformable KPConv).

\begin{table}
\centering
\begin{tabular}{lccc}  
\toprule
Method  & mIoU(\%) \\
\midrule
PointNet ({\color{blue}CVPR'17})~\cite{qi2017pointnet} & 41.09\\
RSNet ({\color{blue}CVPR'18})~\cite{huang2018recurrent} & 51.93\\
PointCNN ({\color{blue}NIPS'18})~\cite{li2018pointcnn} & 57.26\\
ASIS ({\color{blue}CVPR'19})~\cite{wang2019associatively} & 54.48\\
ELGS ({\color{blue}NIPS'19})~\cite{wang2019exploiting} & 60.06\\
PAT ({\color{blue}CVPR'19})~\cite{yang2019modeling} & 60.07\\
SPH3D-GCN ({\color{blue}TPAMI'20})~\cite{lei2020spherical} & 59.5\\
PointASNL ({\color{blue}CVPR'20})~\cite{yan2020pointasnl} & 62.6 \\
FPConv ({\color{blue}CVPR'20})~\cite{lin2020fpconv} & 62.8 \\
Point2Node ({\color{blue}AAAI'20})~\cite{han2019point2node} & 62.96\\
SegGCN ({\color{blue}CVPR'20})~\cite{lei2020seggcn} & 63.6\\
DCM-Net ({\color{blue}CVPR'20})~\cite{Schult_2020_CVPR} & 64.0\\
FusionNet ({\color{blue}ECCV'20})~\cite{zhang2020deep} & 67.2\\
\midrule
RandLA ({\color{blue}CVPR'20})~\cite{hu2020randla} & 62.42\\
RandLA~\cite{hu2020randla} + Ours & 65.09\\
\midrule
KPConv \textit{deform} ({\color{blue}ICCV'19})~\cite{thomas2019kpconv} & 67.1\\
KPConv \textit{deform} + Ours & \textbf{68.73}\\
\bottomrule
\end{tabular}
\caption{Results of indoor scene semantic segmentation on S3DIS Area-5.}
\label{tab:s3dis}
\end{table}

\begin{figure}[ht]
    \centering
    \includegraphics[width=\linewidth]{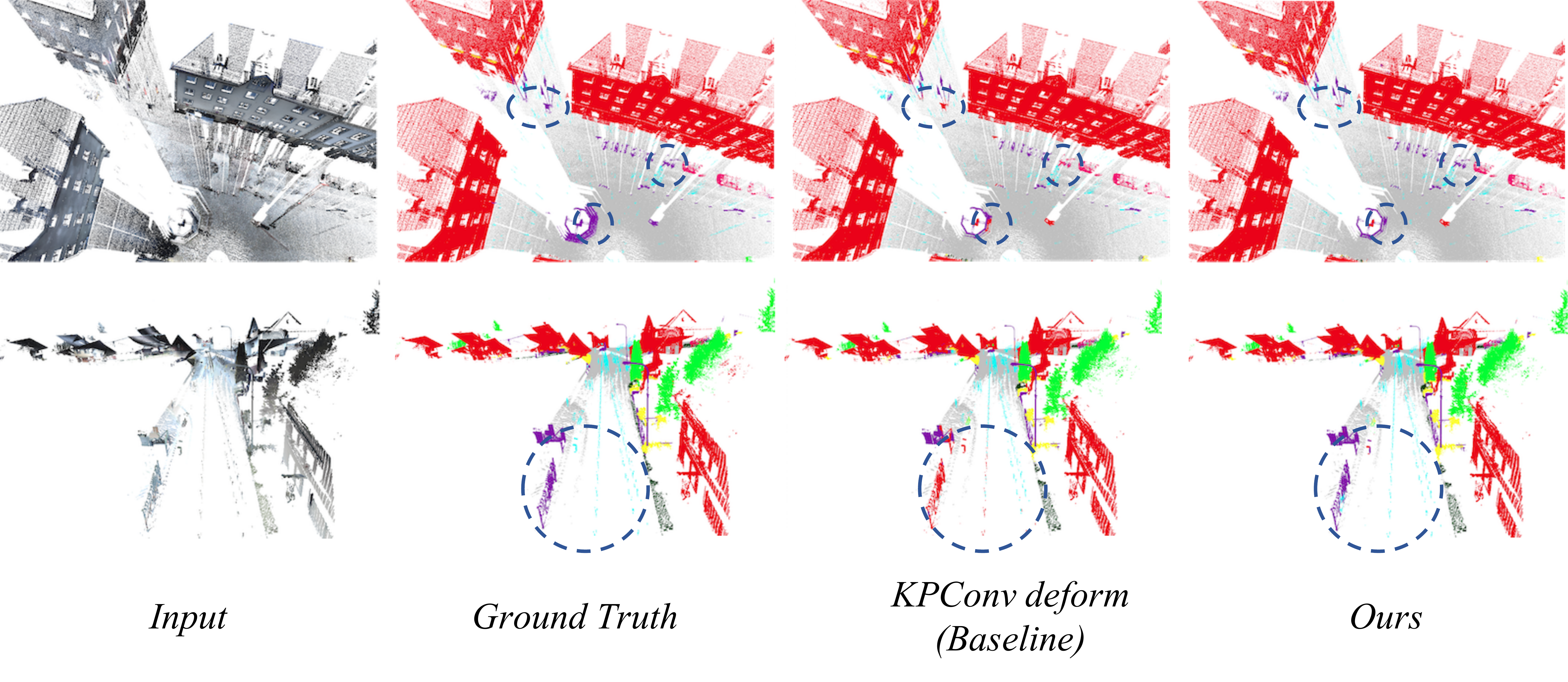}
    \caption{Visualizations on validation set of Semantic3D. Inputs, semantic labels, results of our baseline and our method are presented separately from the left to the right.}
    \label{fig:semantic3d}
\end{figure}

In Table~\ref{tab:semantic3d}, we show the results of our method and other prevailing methods on Semantic3D~\cite{hackel2017semantic3d}. In this task, we achieve $77.8\%$ in mIoU, outperforming all the state-of-the-art competitors. When taking deformable KPConv as our backbone, our method improves it by $4.7\%$. Then we take rigid KPConv as our backbone, and our method can also bring $3.0\%$ improvement in mIoU. We present the visual results of our method and the baseline (deformable KPConv) on the validation set of Semantic3D in Figure~\ref{fig:semantic3d}. The dark blue dashed circles indicate the qualitative improvements.

\begin{table}
\centering
\begin{tabular}{lccc}  
\toprule
Method  & mIoU(\%) \\
\midrule
SegCloud ({\color{blue}3DV'17})~\cite{tchapmi2017segcloud} & 61.3\\
RF\_MSSF ({\color{blue}3DV'18})~\cite{thomas2018semantic} & 62.7\\
SPG ({\color{blue}CVPR'18})~\cite{landrieu2018large} & 73.2\\
ShellNet ({\color{blue}ICCV'19})~\cite{zhang2019shellnet} & 69.4\\
GACNet ({\color{blue}CVPR'19})~\cite{wang2019graph} & 70.8\\
FGCN ({\color{blue}CVPR'20})~\cite{khan2020fgcn} & 62.4\\
PointGCR ({\color{blue}WACV'20})~\cite{ma2020global} & 69.5\\
RandLA ({\color{blue}CVPR'20})~\cite{hu2020randla} & 77.4\\
\midrule
KPConv \textit{rigid} ({\color{blue}ICCV'19})~\cite{thomas2019kpconv} & 74.6\\
KPConv \textit{rigid} + Ours & 77.6\\
\midrule
KPConv \textit{deform} ({\color{blue}ICCV'19})~\cite{thomas2019kpconv} & 73.1\\
KPConv \textit{deform} + Ours & \textbf{77.8}\\
\bottomrule
\end{tabular}
\caption{Results of outdoor space semantic segmentation on Semantic3D (reduced-8).}
\label{tab:semantic3d}
\end{table}

\subsection{Ablation Study}
\label{subsec:ablation}
In this section, we conduct more experiments to evaluate the effectiveness of the proposed gradual Receptive Field Component Reasoning (RFCR) method from different aspects. Without loss of generality, our ablation studies are mainly conducted on the task of Semantic3D reduced-8 and deformable KPConv~\cite{thomas2019kpconv} is chosen as backbone.

\paragraph{Gradual Receptive Field Component Reasoning.}
To conduct ablation studies on different parts of gradual Receptive Field Component Reasoning in the semantic segmentation, we firstly only give the omni-supervision in the decoding procedure to guide the network reason Receptive Field Component Codes (RFCCs) gradually without the loss for Feature Densification (FD). Then, we add the centrifugal potential to obtain more active features for RFCC prediction, and the results are reported in Table~\ref{tab:ablation}. The results indicate the Receptive Field Component Reasoning can improve the segmentation performance by $2.9\%$ alone, and FD can further bring $1.8\%$ improvement. We also conduct ablation studies on the effects of supervisions at different scales and provide the details in supplementary materials.

\begin{table}
\centering
\begin{tabular}{lc} 
\toprule
Method  & mIoU \\
\midrule
KPConv \textit{deform} & 73.1\\
\ \ + RFCR& 76.0\\
\ \ \ \ + FD& 77.8 \\
\bottomrule
\end{tabular}
\caption{Ablation study on impact of different parts of gradual Receptive Field Component Reasoning.}
\label{tab:ablation}
\end{table}

\paragraph{Omni-supervision via Up-sampling.}

Multi-scale supervision is usually used in 2D segmentation via up-sampling the low-resolution prediction. Even we cannot up-sample the point cloud through simple tiling or interpolation, we attempt to up-sample the intermediate predictions iteratively using the nearest neighbors. Then, semantic labels are used to supervise all the up-sampled predictions. Same as our method, all scales are supervised and Feature Densification is also used to provide more unambiguous features for intermediate prediction. We report the result of Omni-supervision via Up-sampling (OvU) in Table~\ref{tab:strategy} and compare it with our method. It shows inferior performance ($76.2\%$) because the up-sampling method using nearest neighbors cannot trace the proper encoding relationship.

\begin{table}
\centering
\begin{tabular}{lc} 
\toprule
Method  & mIoU \\
\midrule
KPConv \textit{deform} & 73.1\\
\midrule
KPConv \textit{deform} + OvU + FD & 76.2\\
KPConv \textit{deform} + RFCR[one-hot] + FD & 76.4\\
\midrule
KPConv \textit{deform} + RFCR + FD& 77.8 \\
\bottomrule
\end{tabular}
\caption{Ablation study on omni-scale supervision strategy.}
\label{tab:strategy}
\end{table}

\paragraph{One-hot RFCC.}
In previous works like PointRend~\cite{kirillov2020pointrend}, they give one-hot predictions at low resolutions, and these predictions will be up-sampled to be supervised by the one-hot labels at original resolution. So, it is intuitive to take an one-hot RFCC for the major category in the receptive field to supervise the prediction. However, the category information of some points will be ignored in this way. Compared with this method, we take a multi-hot label for every sampled point at all the scales, and no labels will be ignored in the supervision of down-sampled points. In order to show the benefit of multi-hot labels, we replace the multi-hot labels with one-hot labels which represent the majority of categories in the receptive fields, and all other settings remain the same. We report the results in Table~\ref{tab:strategy}. We can see one-hot RFCC which ignores the minor category cannot fully represent the information in the receptive field, thus having sub-optimal performance ($76.4\%$) in the segmentation which is $1.4\%$ lower than multi-hot RFCC.


\paragraph{Feature Densification.}

\begin{figure}[ht]
    \centering
    \includegraphics[width=0.85\linewidth]{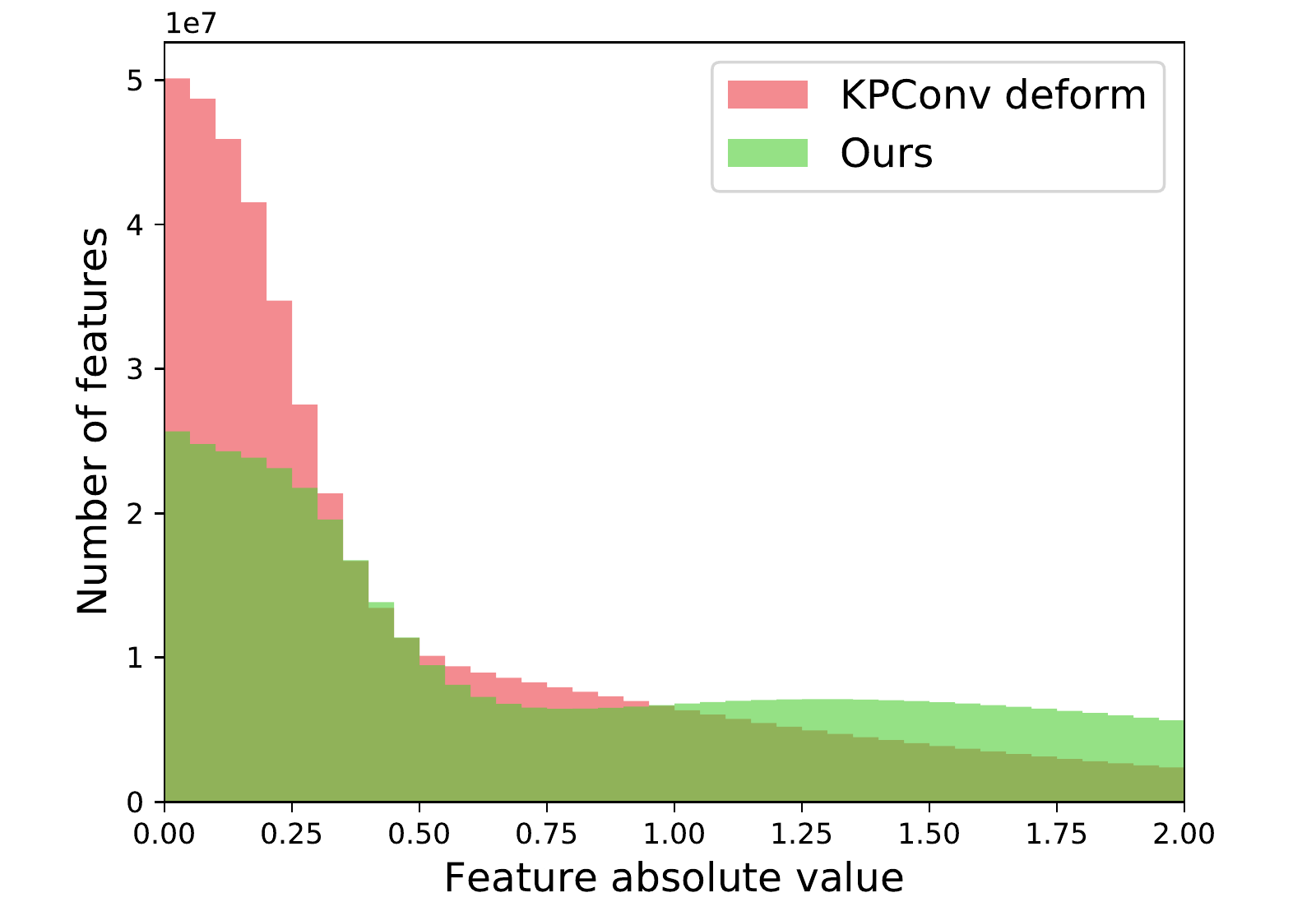}
    \caption{Visualization of features' magnitude in the decoding layers. The green chart bars represent the distribution of features' absolute value after adding Feature Densification while the red chart bars represent the distribution of features' absolute value in the original network.}
    \label{fig:magnitude}
\end{figure}

As stated in Sec~\ref{subsec:fd}, active features will be densified by centrifugal potential given the loss in Eq.~(\ref{eq:fd_loss}). The distribution of features' magnitude after training can be visualized by the bar chart shown in Figure~\ref{fig:magnitude}. As indicated in this figure, features are pushed away from $0$ and more unambiguous features are available for the Receptive Field Component Reasoning, thus improving the segmentation performance (Table~\ref{tab:ablation}).

\section{Conclusion}
In this paper, we propose a gradual Receptive Field Component Reasoning method for omni-supervised point cloud segmentation which decomposes the hard segmentation problem into a global context recognition task and a series of gradual Receptive Field Component Code reasoning steps. Additionally, we propose a complementary Feature Densification method to provide more active features for RFCC prediction. We evaluate our method with four prevailing backbones on three popular benchmarks and outperform almost all the state-of-the-art point-based competitors. Furthermore, our method brings new state-of-the-art performance for Semantic3D and S3DIS benchmarks. Even our method brings large improvements to many backbones for point cloud segmentation, it is more suitable for networks with encoder-decoder architecture.

{\small
\bibliographystyle{ieee_fullname}
\bibliography{cvpr}
}
\newpage
\appendix
\section*{Appendix}

\section{Supervisions at Different Layers}
\label{sec:ablation}

\begin{table}[hb]
\centering
\begin{tabular}{cccccc} 
\toprule
\multicolumn{5}{c}{Supervision Scales} & \multirow{2}{*}{mIoU}\\
\cline{1-5} 
1 & 2 & 3 & 4 & 5  \\
\midrule
\checkmark & \checkmark & \checkmark & \checkmark &  &  76.3\\
\checkmark & \checkmark & \checkmark &  & \checkmark &  76.6\\
\checkmark & \checkmark &  & \checkmark & \checkmark &  76.9\\
\checkmark &  & \checkmark & \checkmark & \checkmark &  76.2\\
\midrule
\checkmark & \checkmark & \checkmark & \checkmark & \checkmark &  77.8 \\
\bottomrule
\end{tabular}
\caption{Ablation study on significance of supervisions at different scales.}
\label{tab:layer}
\end{table}

\begin{figure}
    \centering
    \includegraphics[width=\linewidth]{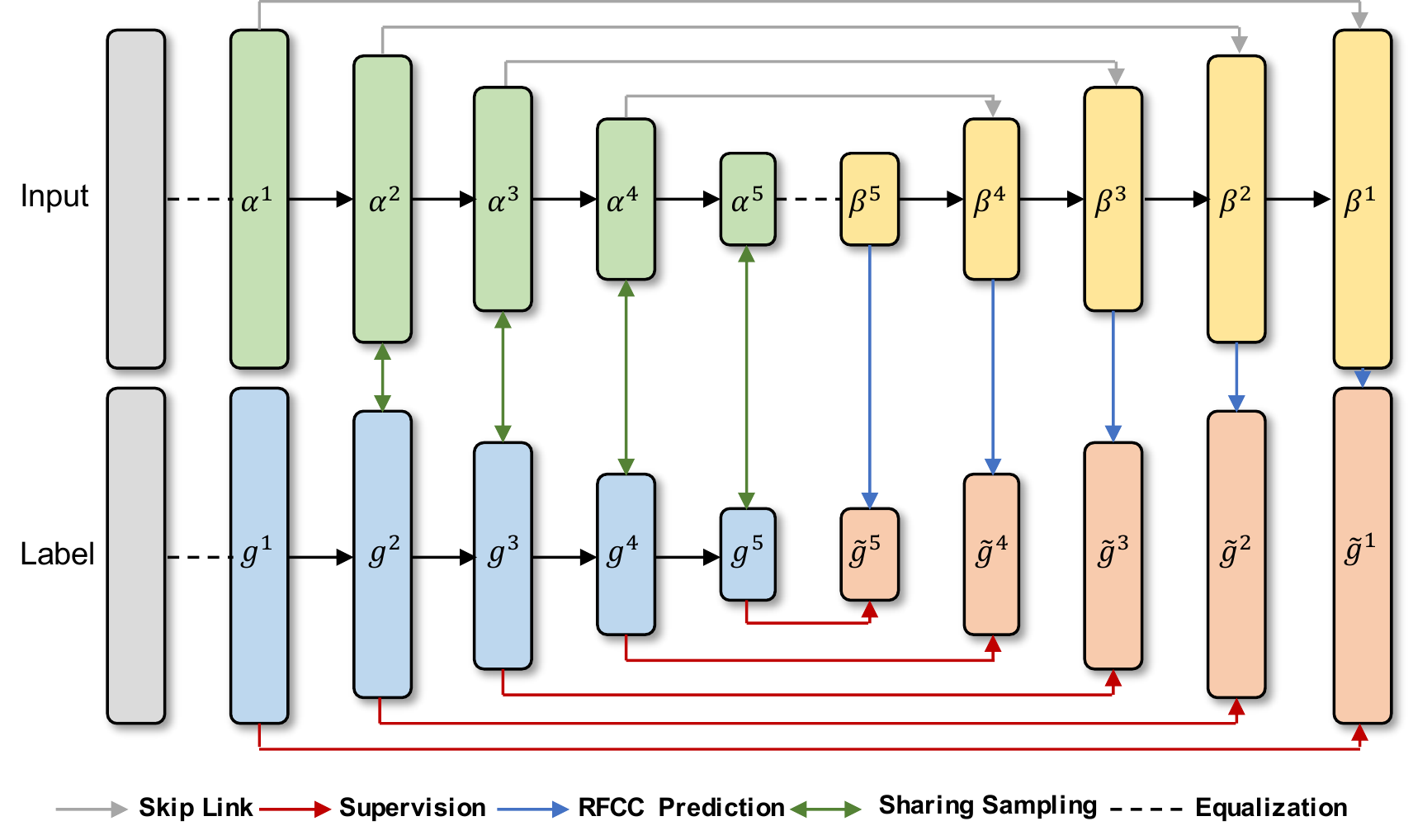}
    \caption{Illustration of framework using deformable KPConv as the backbone. In our method, all the five scales are supervised by the target RFCCs.}
    \label{fig:mini_framework}
\end{figure}

We design an omni-scale supervision method for point cloud segmentation via the proposed gradual Receptive Field Component Reasoning in the main paper. All scales are supervised in the decoding stage to learn informative representation for semantic segmentation. In this section, we attempt to analyze the significance of supervisions at different scales. In this ablation study, deformable KPConv~\cite{thomas2019kpconv} is also taken as the backbone and performance is evaluated on the Semantic3D reduced-8 task. In the architecture of deformable KPConv network, there are 5 different scales as shown in Figure~\ref{fig:mini_framework}. So, we separately remove the supervisions for $l=2,3,4,5$. It is noteworthy that we always keep the supervision for the final layer ($l=1$) because it directly guides the semantic label prediction, otherwise the network will give random prediction. The results is reported in Table~\ref{tab:layer}. The results indicates supervision in the center-most layer ($l=5$) plays an important role in the omni-scale supervision. That is because it can help the encoder to obtain representative global features which is quite important for the following reasoning. Meanwhile, the supervision before the final prediction $l=2$ also contributes a lot because it can directly provide semantic informative features to the final segmentation.

\section{Visualization of intermediate RFCC}
\label{sec:rfcc}
\begin{figure}[h!]
    \centering
    \includegraphics[width=\linewidth]{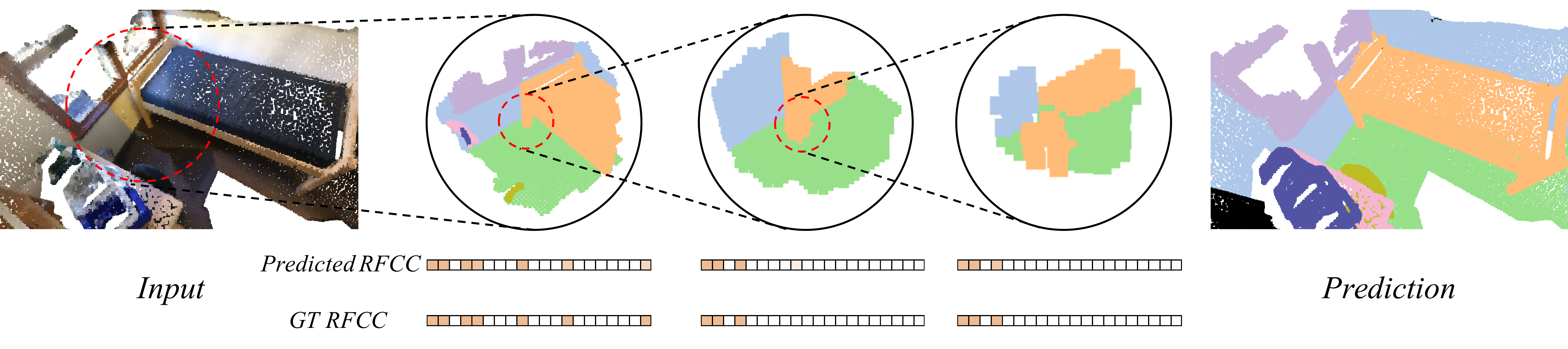}
    \caption{Visualization of intermediate RFCCs whose element color represents the probability of existence for each category. }
    \label{fig:reason}
\end{figure}

We visualize the RFCC reasoning process and our predicted RFCCs in intermediate layers to implicitly show the intermediate feature learning in Figure~\ref{fig:reason}. Meanwhile, the OA of RFCC prediction is $97.34\%$ on the validation set of ScanNet v2, demonstrating good representation learning of intermediate features to some extent.

\section{Supervision on Decoder vs. Encoder}
\label{sec:encoder}

\begin{table}[h]
    \centering
    \begin{tabular}{lc}
    \toprule
        Method &  mIoU\\
    \midrule
        KPConv \textit{deform} & 73.1\\
    \midrule
        KPConv \textit{deform} + [RFCR + FD][encoder] & 76.8\\
        KPConv \textit{deform} + RFCR + FD & 77.8\\
    \bottomrule
    \end{tabular}
    \caption{More ablation study on the strategy of omni-scale supervision.}
    \label{tab:encoder}
\end{table}

In our implementation, all the supervisions are added in the decoder even the target RFCCs are generated according to the receptive fields of features in the encoder. That is because the features in the encoder can also be supervised through the skip links. In order to show the advantage of our strategy, we attempt to supervise the features in the encoder rather than the decoder according to the RFCCs, and Feature Densification is also applied on the corresponding features in the encoder. Compared with supervision in the decoding stage, guiding the feature extraction using RFCCs in the encoder is not able to effectively extract informative representation from global and local features in the decoding stage, such obtaining inferior result as reported in Table~\ref{tab:encoder}. 

\section{Visualization Results}
\label{sec:visual}

\begin{figure}[bht]
    \centering
    \includegraphics[width=\linewidth]{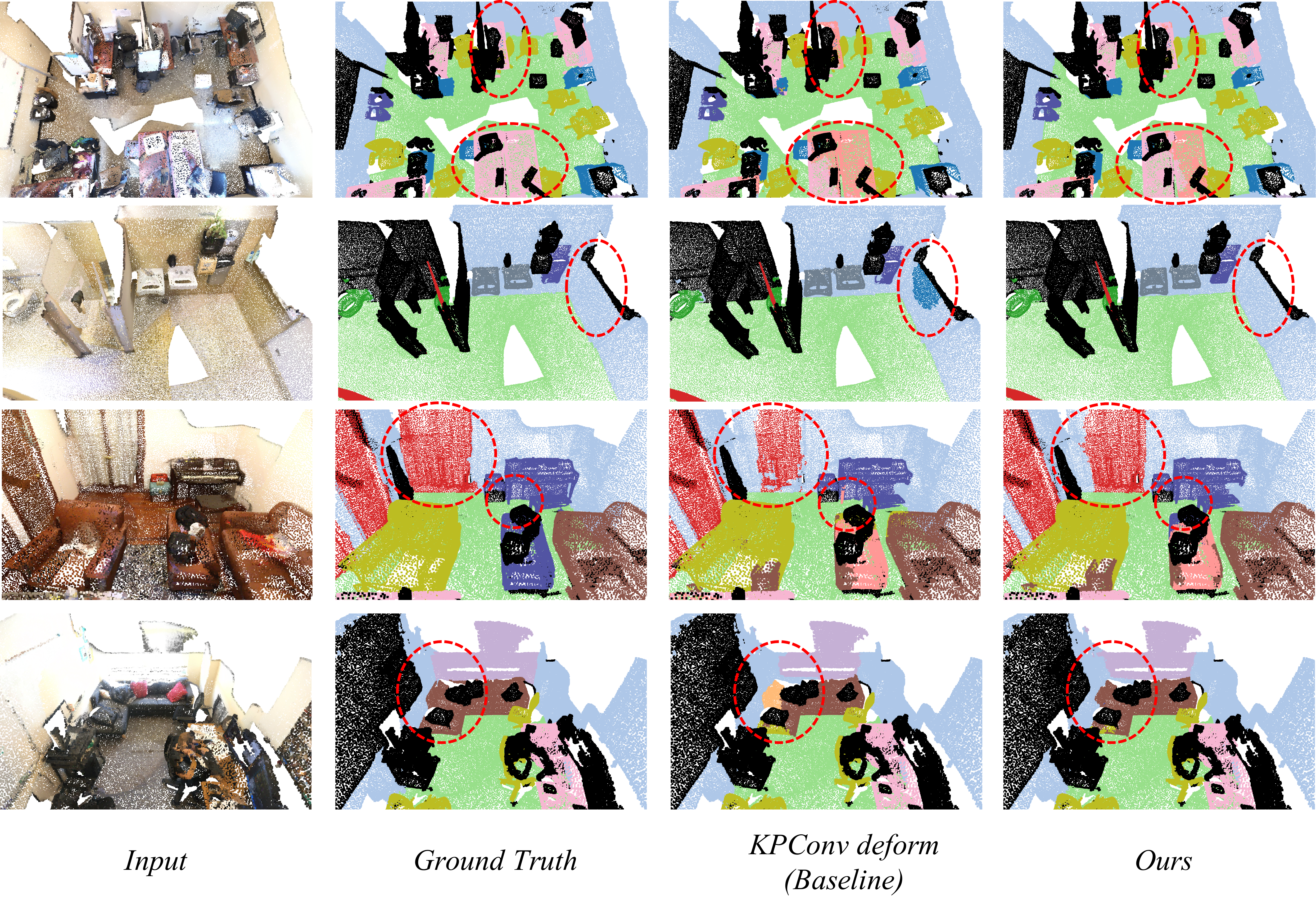}
    \caption{More visualization results on the validation dataset of ScanNet v2. The images from the left to right are input point clouds, semantic labels, predictions given by our baseline and our method, respectively.}
    \label{fig:scannet_supp}
\end{figure}

\begin{figure}[bht]
    \centering
    \includegraphics[width=\linewidth]{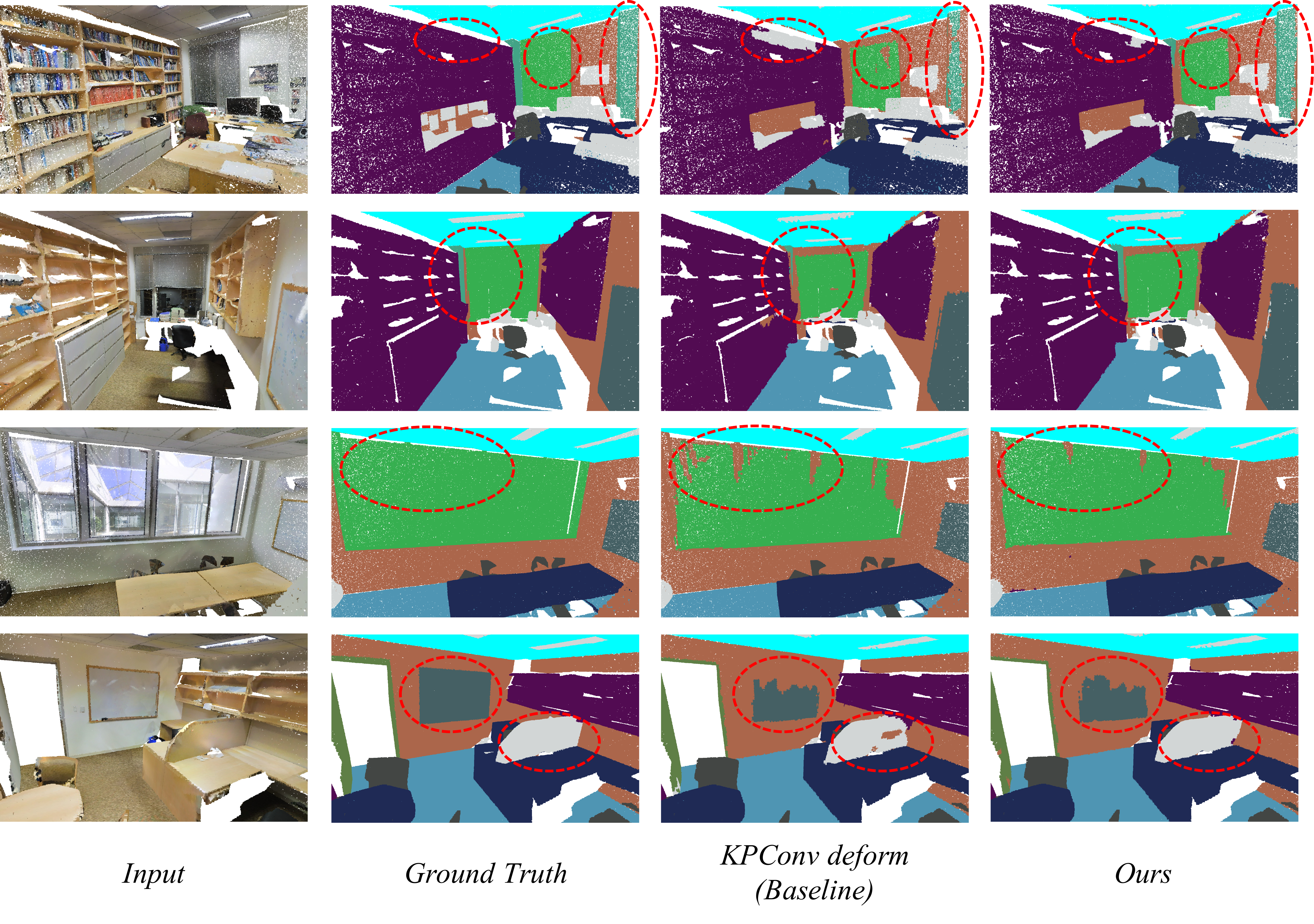}
    \caption{More visualization results on the test dataset of the S3DIS Area-5. The left-most images are inputs and the following images are segmentation ground truth, predictions of baseline and our method separately.}
    \label{fig:s3dis_supp}
\end{figure}

\begin{figure}[bht]
    \centering
    \includegraphics[width=\linewidth]{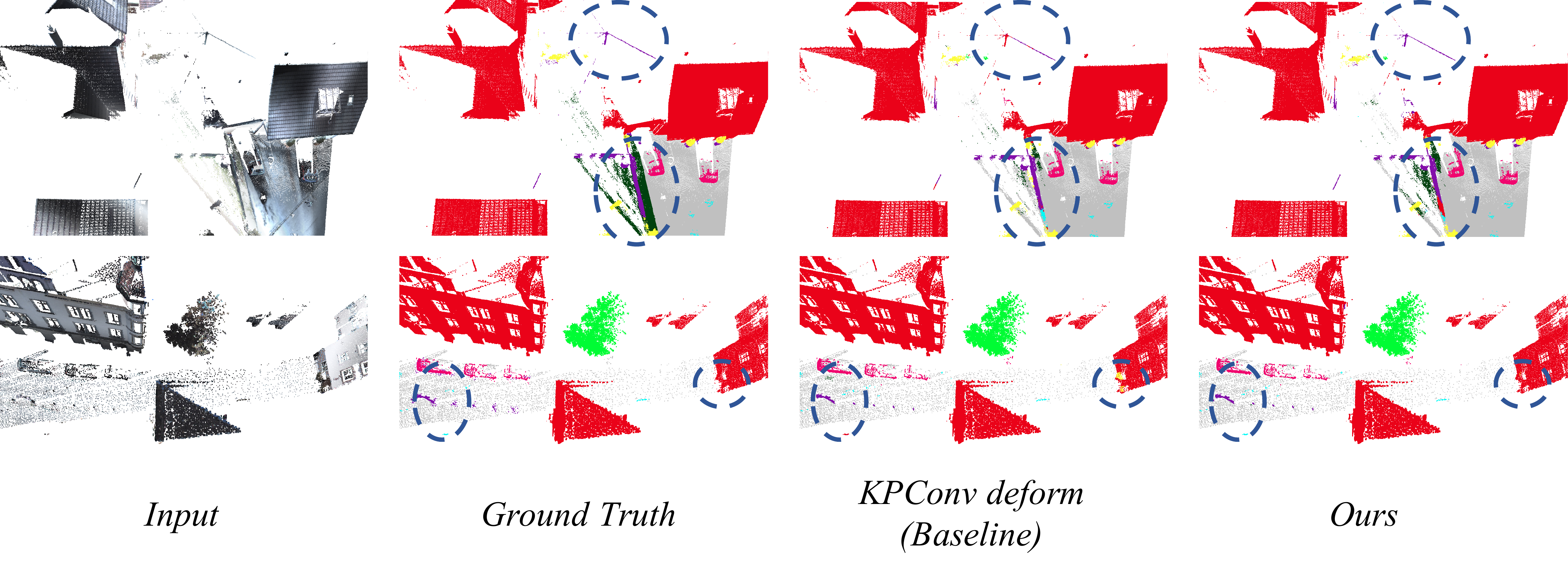}
    \caption{More visualization results on the validation dataset of Semantic3D. Input point clouds, semantic labels, results of our baseline and our method are presented respectively from left to right.}
    \label{fig:semantic3d_supp}
\end{figure}

In this section, we present more visualization results of our method on the three datasets described in the main paper. We present more visualization results of our baseline and our methods on the validation set of ScanNet v2~\cite{dai2017scannet} in Figure~\ref{fig:scannet_supp}. In Figure~\ref{fig:s3dis_supp}, we provide additional visualization results to show the qualitative improvement over the baseline in S3DIS Area 5. We also visualize more scenes in the validation set of Semantic3D in Figure~\ref{fig:semantic3d_supp}.

\renewcommand{\arraystretch}{0.9}
\begin{table*}[tb]
\setlength\tabcolsep{0.5pt}
\begin{footnotesize}
\begin{center}
\begin{tabular}{L{4.0cm} | C{0.7cm} | *{20}{C{0.6cm}}}
\toprule
Method	 & mIoU & bath. & bed & bksf. & cab. & chair & ctr. & curt. & desk & door & floor & oth. & pic. & ref. & shw. & sink & sofa & tab. & toil. & wall & win.	\Bstrut\\
\midrule
PointNet++ ({\color{blue}NIPS'17})~\cite{qi2017pointnet} & 33.9 &58.4	&47.8	&45.8	&25.6	&36.0	&25.0	&24.7	&27.8	&26.1	&67.7	&18.3	&11.7	&21.2	&14.5	&36.4	&34.6	&23.2	&54.8	&52.3	&25.2\\
PointCNN ({\color{blue}NIPS'18})~\cite{li2018pointcnn} & 45.8 &57.7	&61.1	&35.6	&32.1	&71.5	&29.9	&37.6	&32.8	&31.9	&94.4	&28.5	&16.4	&21.6	&22.9	&48.4	&54.5	&45.6	&75.5	&70.9	&47.5\\
3DMV ({\color{blue}ECCV'18})~\cite{dai20183dmv} &48.4 &48.4	&53.8	&64.3	&42.4	&60.6	&31.0	&57.4	&43.3	&37.8	&79.6	&30.1	&21.4	&53.7	&20.8	&47.2	&50.7	&41.3	&69.3	&60.2	&53.9\\
PointConv ({\color{blue}CVPR'19})~\cite{wu2019pointconv} & 55.6& - & - & - & - & - & - & - & - & - & - & - & - & - & - & - & - & - & - & - & -\\
TextureNet ({\color{blue}CVPR'19})~\cite{huang2019texturenet} & 56.6 &67.2	&66.4	&67.1	&49.4	&71.9	&44.5	&67.8	&41.1	&39.6	&93.5	&35.6	&22.5	&41.2	&53.5	&56.5	&63.6	&46.4	&79.4	&68.0	&56.8\\
HPEIN ({\color{blue}ICCV'19})~\cite{jiang2019hierarchical} &61.8 & 72.9	&66.8	&64.7	&59.7	&76.6	&41.4	&68.0	&52.0	&52.5	&94.6	&43.2	&21.5	&49.3	&59.9	&63.8	&61.7	&57.0	&\textbf{89.7}	&80.6	&60.5\\
SegGCN ({\color{blue}CVPR'20})~\cite{lei2020seggcn}  &58.9	&83.3	&73.1	&53.9	&51.4	&78.9	&44.8	&46.7	&57.3	&48.4	&93.6	&39.6	&6.1	&50.1	&50.7	&59.4	&70.0	&56.3	&87.4	&77.1	&49.3\\
SPH3D-GCN ({\color{blue}TPAMI'20})~\cite{lei2020spherical} & 61.0 & 85.8	&77.2	&48.9	&53.2	&79.2	&40.4	&64.3	&57.0	&50.7	&93.5	&41.4	&4.6	&51.0	&70.2	&60.2	&70.5	&54.9	&85.9	&77.3	&53.4\\
    FusionAwareConv ({\color{blue}CVPR'20})~\cite{zhang2020fusion} & 63.0 &60.4	&74.1	&76.6	&59.0	&74.7	&\textbf{50.1}	&73.4	&50.3	&52.7	&91.9	&45.4	&32.3	&55.0	&42.0	&67.8	&68.8	&54.4	&89.6	&79.5	&62.7\\
FPConv ({\color{blue}CVPR'20})~\cite{lin2020fpconv} & 63.9& 78.5	&76.0	&71.3	&60.3	&79.8	&39.2	&53.4	&60.3	&52.4	&94.8	&45.7	&25.0	&53.8	&72.3	&59.8	&69.6	&61.4	&87.2	&79.9	&56.7\\
DCM-Net ({\color{blue}CVPR'20})~\cite{Schult_2020_CVPR} & 65.8 &77.8	&70.2	&80.6	&61.9	&81.3	&46.8	&69.3	&49.4	&52.4	&94.1	&44.9	&29.8	&51.0	&82.1	&67.5	&72.7	&56.8	&82.6	&80.3	&63.7\\
PointASNL ({\color{blue}CVPR'20})~\cite{yan2020pointasnl} & 66.6 & 70.3	&\textbf{78.1}	&75.1	&65.5	&\textbf{83.0}	&47.1	&76.9	&47.4	&53.7	&95.1	&47.5	&27.9	&\textbf{63.5}	&69.8	&67.5	&75.1	&55.3	&81.6	&80.6	&70.3\\
    FusionNet ({\color{blue}ECCV'20})~\cite{zhang2020deep} & 68.8& 70.4 	&74.1 	&75.4 	&65.6 	&82.9 	&\textbf{50.1} 	&74.1 	&60.9 	&54.8 	&95.0	&\textbf{52.2} 	&\textbf{37.1} 	&63.3 	&75.6 	&\textbf{71.5} 	&77.1 	&62.3 	&86.1	&81.4 	&\textbf{65.8} \\
\midrule
SceneEncoder ({\color{blue}IJCAI'20})~\cite{xu2020sceneencoder} & 62.8 & - & - & - & - & - & - & - & - & - & - & - & - & - & - & - & - & - & - & - & -\\
SceneEncoder + Ours & 65.9 &69.1 &72.4 &69.6 &63.2 &81.5 &47.7 &75.4 &\textbf{64.6} &50.9 &\textbf{95.2} &42.8 &28.4 &56.6 &76.1 &62.6 &71.1 &61.0 &88.9 &79.3 &61.0\\
\midrule
KPConv \textit{deform} ({\color{blue}ICCV'19})~\cite{thomas2019kpconv} & 68.4 &84.7	&75.8	&78.4	&64.7	&81.4	&47.3	&77.2	&60.5	&59.4	&93.5	&45.0	&18.1	&58.7	&80.5	&69.0	&\textbf{78.5}	&61.4	&88.2	&81.9	&63.2\\
KPConv \textit{deform} + Ours & \textbf{70.2} & \textbf{88.9} & 74.5 & \textbf{81.3} & \textbf{67.2} & 81.8 & 49.3 & \textbf{81.5} & 62.3 & \textbf{61.0} & 94.7 & 47.0 & 24.9 & 59.4 & \textbf{84.8} & 70.5 & 77.9 & \textbf{64.6} & 89.2 & \textbf{82.3} & 61.1\\
\bottomrule
\end{tabular}
\end{center}
\end{footnotesize}
\caption{Semantic segmentation results on ScanNet v2.}
\label{tab:scannet_supp} 
\end{table*}

\renewcommand{\arraystretch}{0.9}
\begin{table*}[thb]
\setlength\tabcolsep{0.5pt}
\begin{footnotesize}
\begin{center}
\begin{tabular}{L{4.0cm} | C{1.0cm} | *{13}{C{0.92cm}}}
\toprule
Method	 & mIoU & ceil.	 & floor	 & wall	 & beam	 & col.	 & wind.	 & door	 & chair	 & table	 & book.	 & sofa	 & board & clut.	\Bstrut\\
\midrule
PointNet ({\color{blue}CVPR'17})~\cite{qi2017pointnet} & 41.09 & 88.80 & 97.33 & 69.80 & 0.05 & 3.92 & 46.26 & 10.76 & 58.93 & 52.61 & 5.85 & 40.28 & 26.38 & 33.22\\
RSNet ({\color{blue}CVPR'18})~\cite{huang2018recurrent} & 51.93 & 93.34 & 98.36 & 79.18 & 0.00 & 15.75 & 45.37 & 50.10 & 65.52 & 67.87 & 22.45 & 52.45 & 41.02 & 43.64\\
PointCNN ({\color{blue}NIPS'18})~\cite{li2018pointcnn} & 57.26 & 92.31 & 98.24 & 79.41 & 0.00 & 17.6 & 22.77 & 62.09 & 74.39 & 80.59 & 31.67 & 66.67 & 62.05 & 56.74\\
ASIS ({\color{blue}CVPR'19})~\cite{wang2019associatively} & 53.40 & - & - & - & - & - & - & - & - & - & - & - & - & -\\
ELGS ({\color{blue}NIPS'19})~\cite{wang2019exploiting} & 60.06 & 92.80 & 98.48 & 72.65 & 0.01 & 32.42 & 68.12 & 28.79 & 74.91 & 85.12 & 55.89 & 64.93 & 47.74 & 58.22\\
PAT ({\color{blue}CVPR'19})~\cite{yang2019modeling}&60.07 & 93.04 & 98.51 & 72.28 & \textbf{1.00} & \textbf{41.52} & \textbf{85.05} & 38.22 & 57.66 & 83.64 & 48.12 & 67.00 & 61.28 & 33.64 \\
SPH3D-GCN ({\color{blue}TPAMI'20})~\cite{lei2020spherical} & 59.5 & 93.3 & 97.1 & 81.1 & 0.0 & 33.2 & 45.8 & 43.8 & 79.7 & 86.9 & 33.2 & 71.5 & 54.1 & 53.7\\
PointASNL ({\color{blue}CVPR'20})~\cite{yan2020pointasnl} & 62.6 & 94.3 & 98.4 & 79.1 & 0.0 & 26.7 & 55.2 & 66.2 & 83.3 & 86.8 & 47.6 & 68.3 & 56.4 & 52.1\\
FPConv ({\color{blue}CVPR'20})~\cite{lin2020fpconv} & 62.8 & \textbf{94.6} & 98.5 & 80.9 & 0.0 & 19.1 & 60.1 & 48.9 & 80.6 & 88.0 & 53.2 & 68.4 & 68.2 & 54.9\\
Point2Node ({\color{blue}AAAI'20})~\cite{han2019point2node} & 62.96 & 93.88 & 98.26 & 83.30 & 0.00 & 35.65 & 55.31 & 58.78 & 79.51 & 84.67 & 44.07 & 71.13 & 58.72 & 55.17\\
SegGCN ({\color{blue}CVPR'20})~\cite{lei2020seggcn} & 63.6 & 93.7 & \textbf{98.6} & 80.6 & 0.0 & 28.5 & 42.6 & \textbf{74.5} & 80.9 & \textbf{88.7} & 69.0 & 71.3 & 44.4 & 54.3\\
DCM-Net ({\color{blue}CVPR'20})~\cite{Schult_2020_CVPR} & 64.0 & 92.1 & 96.8 & 78.6 & 0.0 & 21.6 & 61.7 & 54.6 & 78.9 & \textbf{88.7} & 68.1 & 72.3 & 66.5 & 52.4 \Bstrut\\
FusionNet ({\color{blue}ECCV'20})~\cite{zhang2020deep} & 67.2 & - & - & - & - & - & - & - & - & - & - & - & - & -\\
\midrule
RandLA ({\color{blue}CVPR'20})~\cite{hu2020randla}	& 62.42	& 91.19	& 95.66	& 80.11	& 0.00	& 25.24	& 62.27	& 47.36	& 75.78	& 83.17	& 60.82	& 70.82	& 65.15	& 53.95\\
RandLA+Ours & 65.09 & 92.66 & 97.43 & 82.40 & 0.00 & 37.04 & 59.72 & 52.30 & 77.49 & 86.95 & 63.48 & 71.99 & 70.54 & 54.13 \Bstrut\\
\midrule
KPConv \textit{deform} ({\color{blue}ICCV'19})~\cite{thomas2019kpconv}	& 67.1	& 92.8	& 97.3	& 82.4	& 0.0	& 23.9	& 58.0	& 69.0	& 91.0	& 81.5	& 75.3	& \textbf{75.4}	& 66.7	& 58.9\\
KPConv \textit{deform}+Ours & \textbf{68.73} & 94.18 & 98.33 & \textbf{84.34} & 0.00 & 28.45 & 62.36 & 71.17 & \textbf{91.95} & 82.60 & \textbf{76.13} & 71.14 & \textbf{71.60} & \textbf{61.25} \Bstrut\\
\bottomrule
\end{tabular}
\end{center}
\end{footnotesize}
\caption{Results of indoor scene semantic segmentation on S3DIS Area-5.}
\label{tab:s3dis_supp} 
\end{table*}

\renewcommand{\arraystretch}{0.9}
\begin{table*}[!h]
\setlength\tabcolsep{0.5pt}
\begin{footnotesize}
\begin{center}
\begin{tabular}{L{4.2cm} | C{1.cm} | *{8}{C{1.5cm}}}
\toprule
Method	 & mIoU & man-made.	 & natural.	 & high veg. & low veg.	 & buildings	 & hard scape & scanning. & cars	\Bstrut\\
\midrule
SegCloud ({\color{blue}3DV'17})~\cite{tchapmi2017segcloud} & 61.3 & 83.9 & 66.0 & 86.0 & 40.5 & 91.1 & 30.9 & 27.5 & 64.3\\
RF\_MSSF ({\color{blue}3DV'18})~\cite{thomas2018semantic} & 62.7& 87.6 & 80.3 & 81.8 & 36.4 & 92.2 & 24.1 & 42.6 & 56.6\\
SPG ({\color{blue}CVPR'18})~\cite{landrieu2018large} & 73.2& 97.4& \textbf{92.6} & 87.9& 44.0 & 93.2 & 31.0 & 63.5 & 76.2\\
ShellNet ({\color{blue}ICCV'19})~\cite{zhang2019shellnet} & 69.4 & 96.3 & 90.4 & 83.9 & 41.0 & 94.2 & 34.7 & 43.9 & 70.2\\
GACNet ({\color{blue}CVPR'19})~\cite{wang2019graph} & 70.8& 86.4 & 77.7 & \textbf{88.5} & 60.6 & 94.2 & 37.3 & 43.5 & 77.8\\
FGCN ({\color{blue}CVPR'20})~\cite{khan2020fgcn} & 62.4& 90.3 & 65.2 & 86.2 & 38.7 & 90.1 & 31.6 & 28.8 & 68.2\\
PointGCR ({\color{blue}WACV'20})~\cite{ma2020global} & 69.5& 93.8 & 80.0 & 64.4 & \textbf{66.4} & 93.2 & 39.2 & 34.3 & \textbf{85.3}\\
RandLA ({\color{blue}CVPR'20})~\cite{hu2020randla} & 77.4& 95.6 & 91.4 & 86.6 & 51.5 & \textbf{95.7} & \textbf{51.5} & 69.8 & 76.8\\
\midrule
KPConv \textit{rigid} ({\color{blue}ICCV'19})~\cite{thomas2019kpconv}	& 74.6	&  90.9 & 82.2 & 84.2 & 47.9 & 94.9 & 40.0 & 77.3 & 79.7\\
KPConv \textit{deform} + Ours & 77.6 & \textbf{97.0} & 90.9 & 86.7	& 50.8	& 94.5	& 37.3	& \textbf{79.7}	& 84.1\Bstrut\\
\midrule
KPConv \textit{deform} ({\color{blue}ICCV'19})~\cite{thomas2019kpconv}	& 73.1	&  - & - & - & - & - & - & - & -\\
KPConv \textit{deform} + Ours & \textbf{77.8} & 94.2 & 89.1 & 85.7 & 54.4 & 95.0 & 43.8 & 76.2 & 83.7 \Bstrut\\
\bottomrule
\end{tabular}
\end{center}
\end{footnotesize}
\caption{Semantic segmentation results on Semantic3D (reduced-8).}
\label{tab:semantic3d_supp} 
\end{table*}

\section{Detailed Experimental Results}
\label{sec:experiment}

In this section, we provide more quantitative details about our experimental results for better comparison with other competitors. In Table~\ref{tab:scannet_supp}, we present the mean IoU (mIoU) over categories and the IoUs for different classes for ScanNet v2. We also list the category scores for S3DIS Area-5 in Table~\ref{tab:s3dis_supp}. It's noteworthy that all the methods do not have good performance on the segmentation of beams in Area 5 because there is a large difference between the beams in Area 5 (test set) and those in Area 1, 2, 3, 4, and 6 (training set). Finally, Table~\ref{tab:semantic3d_supp} shows the IoUs of various classes for Semantic3D reduced-8 task.

\end{document}